\documentclass[11pt]{article}
\usepackage{geometry}
\usepackage{coling2020}
\usepackage{times}
\usepackage{url}
\usepackage{latexsym}
\usepackage{microtype}
\usepackage{tikz}
\usepackage{caption}
\usetikzlibrary{arrows,positioning} 
\tikzset{
    >=stealth',
    word/.style={
           rectangle,
           rounded corners,
           draw=black, very thick,
           line width=0.3mm,
           text width=1.5em,
           scale=0.8,
           minimum height=2em,
           text centered},
    layer/.style={
           rectangle,
           rounded corners,
           draw=black, very thick,
           line width=0.3mm,
           text width=15em,
           minimum height=2.5em,
           text centered},
    pil/.style={
           ->,
           thick,
           shorten <=2pt,
           shorten >=2pt,}
}

\colingfinalcopy % Uncomment this line for all SemEval submissions

\title{Palomino-Ochoa at SemEval-2020 Task 9: Robust System based on Transformer for Code-Mixed Sentiment Classification}

\author{Daniel Palomino \\
Dept. of Computer Science\\
  Universidad Católica San Pablo \\ Arequipa, Peru\\
  {\tt daniel.palomino.paucar@ucsp.edu.pe} \\\And
  José Ochoa-Luna \\
  Dept. of Computer Science\\
  Universidad Católica San Pablo \\
    Arequipa, Peru\\
  {\tt jeochoa@ucsp.edu.pe} \\}

\date{}

\begin{document}
\maketitle
\begin{abstract}
We present a transfer learning system to perform a mixed Spanish-English sentiment classification task. Our proposal uses the state-of-the-art language model BERT and embed it within a ULMFiT transfer learning pipeline. This combination allows us to predict the polarity detection of code-mixed (English-Spanish) tweets. Thus, among 29 submitted systems, our approach (referred to as \textbf{dplominop}) is ranked 4th on the Sentimix Spanglish test set of SemEval 2020 Task 9. In fact, our system yields the weighted-F1 score value of 0.755 which can be easily reproduced --- the source code and implementation details are made available.

%We present a supervised sentiment .. which was used to compete in SemEval 2020 Task 9  Sentimix: Sentiment Analysis for Code-Mixed Social Media Text. 

%In this paper we report our work for SemEVal 2020 task 9 challenge. Our submisison ias based on Deep Learning,  We ... We . The The system yields the F-score value of .  
  
\end{abstract}

% Oficial Guidelines https://semeval.github.io/system-paper-template.html
\section{Introduction}
\label{introduction}

%
% The following footnote without marker is needed for the camera-ready
% version of the paper.
% Comment out the instructions (first text) and uncomment the 8 lines
% under "final paper" for your variant of English.
% 
\blfootnote{
    %
    % for review submission
    %
    % \hspace{-0.65cm}  % space normally used by the marker
    % Place licence statement here for the camera-ready version. See
    % Section~\ref{licence} of the instructions for preparing a
    % manuscript.
    %
    % % final paper: en-uk version 
    %
    % \hspace{-0.65cm}  % space normally used by the marker
    % This work is licensed under a Creative Commons 
    % Attribution 4.0 International Licence.
    % Licence details:
    % \url{http://creativecommons.org/licenses/by/4.0/}.
    % 
    % final paper: en-us version 
    
    \hspace{-0.45cm}  % space normally used by the marker
    This work is licensed under a Creative Commons 
    Attribution 4.0 International License.
    License details:
    \url{http://creativecommons.org/licenses/by/4.0/}.
}

Sentiment Analysis is one of the most active research areas in Natural Language Processing (NLP), web mining and social media analytics \cite{Tang2016}. In particular, its polarity detection task has been extensively researched since 2002 \cite{Liu2012}.  Polarity detection involves to determine whether a given text, containing an opinion, is positive, negative or neutral. 
In order to solve this task, several dictionary-based methods were proposed in the past \cite{Liu2012}. However, Machine Learning (ML) approaches have been the ones with more impact regarding the-state-of-the-art \cite{Liu2015}. Moreover, traditional ML approaches based on feature engineering and modelling have been consistently improved by Deep Learning methods \cite{Zhang2018}.

As a rule, to apply Deep Learning in sentiment analysis input text should be encoded in some way, for instance using word embeddings. Word embeddings are useful because allow us to encode semantic similarities among words. One can obtain a word vector representation by training a large corpus using algorithms such as Word2vec \cite{w2vec}, Glove \cite{glove2014} and FastText \cite{FastText2017}, to name a few.

Nowadays, this kind of text representation has evolved to a language model encoding. The idea is to use language context in order to better encode words and characters. Overall, the aim of this encoding is to transfer the knowledge embodied in the language model to address a specific task.

Thus, in this paper we focus on transfer learning: we make use of a pre-trained language model and then we applied it in sentiment classification (The aim is to predict the correct sentiment classification of a given code-mixed tweet). In order to do so, our work combines two powerful frameworks: the Universal Language Model Fine-tuning (ULMFiT) \cite{Howard:2018} and BERT \cite{Devlin:2018}. 

ULMFiT is the basis for our transfer learning strategy and it has proven to have an impressive performance on several English text classification tasks---It also has obtained optimal results for the Spanish language \cite{Palomino:2019}. On the other hand, BERT is currently the-state-of-the-art language model for NLP. Hence, we use a reduced multilingual language model (BERT) which is further fine-tuned (ULMFiT) in order to classify code-mixed tweets.

We evaluate our system on the Sentimix Spanglish test set of SemEval 2020 Task 9. Our submission (referred to as \textbf{dpalominop}) obtained the weighted-F1 value of 0.755 and was ranked 4th among 29 teams. 
% It is worth noting that those results could be reproduced because the source code is made available \footnote{https://github.com/dpalominop/atlas}.

% \footnote{https://ritual-uh.github.io/sentimix2020/}
\section{Background}
\label{background}

The aim of the task 9 \cite{patwa2020sentimix} is to predict the correct sentiment classification of a given code-mixed tweet. Each sentiment label can be one of three options: positive, negative and neutral. 

There are two code-mixed languages provided on this track: English-Hindi and English-Spanish. In this work we are only approached the code-mixed English-Spanish task.

Train, evaluation and test datasets are available and contain tweets in CoNLL-U  format \cite{Buchholz:2006}. Thus, every tweet word is tagged accordingly: en (English), spa (Spanish), hi (Hindi), mixed and univ (e.g. symbols, @ mentions, hashtags, etc). The whole tweet is tagged with the corresponding sentiment label.  During the competition only train and evaluation datasets were labeled.

\section{System Overview}
\label{system_overview}

In this section, we present the system design choices that allow us to predict the sentiment of a given code-mixed tweet. 

Overall, our strategy is as follows, due to a small dataset is provided we plan to use a pre-trained language model. By doing so, we aim at extracting features and context words from a large corpus in order to ``transfer'' this knowledge to our small dataset. Consequently, we only should perform a fine-tuning step regarding the task at hand.

Several language models have been proposed in the last years \cite{Mccann2017,Peters2018,Howard:2018}, but the one with the highest impact in the-state-of-the-art has been BERT \cite{Devlin:2018}. BERT is a powerful language model that uses a bidirectional representation from unlabeled data by jointly conditioning on both left and right context words. In order to train this model, Devlin et al. \shortcite{Devlin:2018} propose a novel training task called Masked Language Model (MLM). In this task, some input tokens are randomly masked in order to predict vocabulary identifiers using context, i.e., remaining words of the sentence. In addition, the work proposes the Next Sentence Prediction (NSP) training task that jointly pre-trains text-pairs representations. To do so, the whole text is parsed in several batches of two consecutive phrases. Then, the first one is used to predict the second one. With these two training tasks, BERT has outperformed previous state of the art results on several NLP challenges.

Transfer learning approaches based on language models for NLP tasks have been proposed in the past \cite{Peters2018,Howard:2018,Devlin:2018}. Arguably, the simplest proposal for performing text classification using language models is the Universal Language Model Fine-tuning (ULMFiT) \cite{Howard:2018}. Such work was extended to Spanish sentiment analysis with remarkable results \cite{Palomino:2019}.   

\begin{figure}[h]

\centering
\captionsetup{justification=centering}
\begin{tikzpicture}[node distance=1cm, auto,]

% second part
\node[layer] (classification) {Classification Layer};
\node[word, inner sep=5pt,below=0.5cm of classification] (o22) {$O_{2}$}
edge[->,thick] (classification.south);
\node[word, left=of o22] (o12) {$O_{1}$};
\node[word, right=of o22] (o32) {$O_{3}$};

\node[layer, below=0.5 of o22] (encoder2) {BERT Encoder}
edge[->,thick] (o22.south);
\node[word, inner sep=5pt,below=0.5cm of encoder2] (w22) {$W_{2}$}
edge[->,thick] (encoder2.south);
\node[word, left=of w22] (w12) {$W_{1}$};
\node[word, right=of w22] (w32) {$W_{3}$};

\path[pil, ->]
    (classification.east) edge[bend left] node [pil, above, rotate=270] {ULMFiT fine-tuning} (encoder2.east);

\draw[->,thick] (o12.north) -- (o12.north |- classification.south);
\draw[->,thick] (o32.north) -- (o32.north |- classification.south);

\draw[->,thick] (o12.south |- encoder2.north) -- (o12.south);
\draw[->,thick] (o32.south |- encoder2.north) -- (o32.south);

\draw[->,thick] (w12.north) -- (w12.north |- encoder2.south);
\draw[->,thick] (w32.north) -- (w32.north |- encoder2.south);

% first part
\node[layer, left=of classification] (task) {Pre-training Task Layer}
edge[pil,->, bend left=25] node[auto] {Swapping last layer} (classification);
\node[word, inner sep=5pt,below=0.5cm of task] (o2) {$O_{2}$}
edge[->,thick] (task.south);
\node[word, left=of o2] (o1) {$O_{1}$};
\node[word, right=of o2] (o3) {$O_{3}$};

\node[layer, below=0.5 of o2] (encoder) {BERT Encoder}
edge[->,thick] (o2.south);
\node[word, inner sep=5pt,below=0.5cm of encoder] (w2) {$W_{2}$}
edge[->,thick] (encoder.south);
\node[word, left=of w2] (w1) {$W_{1}$};
\node[word, right=of w2] (w3) {$W_{3}$};

\draw[->,thick] (o1.north) -- (o1.north |- task.south);
\draw[->,thick] (o3.north) -- (o3.north |- task.south);

\draw[->,thick] (o1.south |- encoder.north) -- (o1.south);
\draw[->,thick] (o3.south |- encoder.north) -- (o3.south);

\draw[->,thick] (w1.north) -- (w1.north |- encoder.south);
\draw[->,thick] (w3.north) -- (w3.north |- encoder.south);
        
\end{tikzpicture}

\caption{General system pipeline. $W_{x}$ is the input to the system and \\$O_{x}$ is the output before the pre-training task layer (MLM / NSP).}
\label{pipeline}
\end{figure}
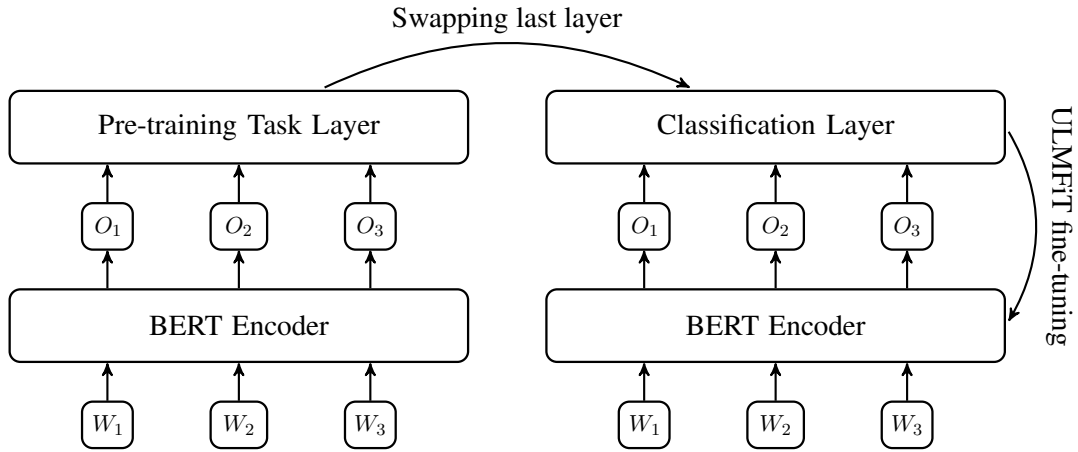

Our proposal uses a base BERT language model and embed it within a ULMFiT transfer learning pipeline. The resulting system is depicted in Figure \ref{pipeline}.

The components are described as follows:

\begin{enumerate}
    \item BERT is the base language model (LM) which is trained on a general domain corpus to capture general features of the language through several layers. Regarding the original ULMFiT pipeline, the LSTM-based LM has been changed by a BERT LM.  
    
    \item The last layer used in BERT to perform NSP pre-training is changed by a single classification layer with the same number of outputs as labels provided on the SEMEVAL challenge \cite{patwa2020sentimix}.

    \item Fine-tuning is performed as in the original ULMFiT pipeline. The target task (sentiment analysis) is tuned through gradual unfreezing, discriminative fine-tuning (Discr), and slanted triangular learning rates (STLR) \cite{Howard:2018}. The aim is to preserve low-level representations and adapt to high-level ones.
    In our context, the sentiment analysis classifier is fine-tuned using the provided labeled English-Spanish tweets. 
\end{enumerate}

So far, we have presented the main pipeline of our system, but research on BERT models have produced several variants such as BASE, LARGE and MULTILINGUAL which rely on the number of parameters, layers and languages, accordingly. As discussed in the original paper \cite{Devlin:2018}, the best choice to use in small and medium datasets is the BASE version because the lower number of parameters to fine-tune. 

However, the code-mixed tweets aimed to classify are written in English and Spanish. In this sense, the best choice would be a multilingual version but,  experimental results have shown that a multilingual LM performs worst than a single LM \footnote{https://github.com/google-research/bert/blob/master/multilingual.md}.

Regarding those results, our design choice has been to use a multilingual model with few languages (including at least English and Spanish) which we referred to as \textbf{reduced multilingual}.

In the next section we describe the pre-processing data, system configuration as well as the choice of the language model used in this challenge.

\section{Experimental Setup}
\label{experimental_setup}

A complete description about hardware and software requirements for reproducing this paper is detailed in this section. In addition, we show the hyper-parameters tuned during experimentation, e.g.  the learning rate that allow us to converge without overfitting and regularization.

\subsection{Technical Resources}

All experiments were carried out on Jupyter notebooks running Python 3.7 kernel and PyTorch 1.3.1. For a detailed explanation about dependencies, please refer to the public project repository \footnote{https://github.com/dpalominop/atlas}.

\subsection{Datasets}

The data is splitted as follows. The training dataset: 12002 labeled tweets, the validation dataset: 2998 labeled tweets and the testing dataset: 3789 unlabeled tweets. Each sentiment label can be one of three options: positive, negative and neutral.

\subsection{Pre-Processing}
All the datasets were pre-processed according to the following rules:

\begin{enumerate}
    \item Every tweet structure was converted to plain text and tagging each word.
    \item Text was converted to lowercase and every accent mark was removed.
    \item Repeated characters were replaced to single characters.
     \item User references, hashtags and useless spaces were removed. 
\end{enumerate}

\subsection{Pre-trained Language Model}

In order to accomplish the constraints presented in section \ref{system_overview}, the pre-trained language model used was  \textbf{bert-base-multilingual-uncased-sentiment}\footnote{https://huggingface.co/nlptown/bert-base-multilingual-uncased-sentiment} which was published in public official repository of Huggingface Co.

\subsection{Fine-tuned Language Model}

The main hyper-parameters used through the fine-tuning process are:

\begin{enumerate}
    \item Backpropagation Trough Time (BPTT): 70
    \item Weight Decay (WD): $1e-2$
    \item The batch size (BS) was limited by the available GPU memory. In our case: 16
\end{enumerate}

\section{Results}
\label{results}

Results for SemEval 2020 Task 9 Competition are reported in Table \ref{top_results}. Our submission (referred to as {\bf dpalominop}) was ranked 4th among 29 systems (weighted-F1 score).

\begin{table}[h]
\begin{center}
\begin{tabular}{l|ccc}
\hline \bf Team & \bf Score 1 (Best Score) & \bf Score 2 & \bf Score 3 \\ \hline
LiangZhao & 0.806 & 0.805 & 0.794 \\
rachel & 0.776 & 0.755 & 0.749 \\
asking28 & 0.756 & 0.612 & 0.595 \\
\bf dpalominop & 0.755 & 0.742 & 0.703 \\
kongjun & 0.753 & 0.726 & 0 \\
\hline
\end{tabular}
\end{center}
\caption{Top 5 results on SemEval 2020 Task 9 test dataset (weighted-F1 Score).}
\label{top_results}
\end{table}

Several techniques were tested before finding our solution. Since we have used a transfer learning approach, one first challenge was to find out the language model that best fit our multilingual constraints. After a thorough analysis, we decided to use a multilingual language model (bert-base-multilingual-uncased-sentiment). This model employs a small number of languages in its pre-training process. Furthermore, several pre-processing data and fine-tuning techniques were also tested.  

We have also performed some ablation experiments regarding our proposal steps so as to understand the impact of the design choices in our results. This analysis is presented in Table \ref{ablation}. In short, although data pre-processing and fine-tuning increase performance, the language model choice have achieved the greatest results.

\begin{table}[h]
\begin{center}
\begin{tabular}{lr}
    \hline
         &   \textbf{Weighted-F1}\\
    \hline
    Our proposal\\
    \hspace{3mm}Using reduced multilingual $BERT_{base}$  &  0.695\\
    \hspace{3mm}w/pre-processing data  &  0.713\\
    \hspace{3mm}w/pre-processing data + ulmfit fine-tuning & \textbf{0.755}\\
    \hline
    Changing pre-trained language model\\
    \hspace{3mm}Using Google multilingual $BERT_{base}$  &  0.613\\
    \hspace{3mm}w/pre-processing data  &  0.636\\
    \hspace{3mm}w/pre-processing data + ulmfit fine-tuning & \textbf{0.681}\\
    \hline
    \end{tabular}
\end{center}
\caption{Ablation using different pre-trained language models. "w/" denotes "with".}
\label{ablation}
\end{table}

\section{Conclusion}
\label{conclusion}

We have presented a transfer learning approach to tackle a mixed Spanish-English sentiment classification task. In order to do so, we have combined a transfer learning scheme based on ULMFiT with the-state-of-the-art language model BERT. That approach allowed us to be ranked 4th on the Sentimix Spanglish test set of SemEval 2020 Task 9.

Furthermore, we have demonstrated that a reduced multilingual language model performs better than the one supporting several languages. Moreover, we have discussed the impact of a correct fine-tuning using a discriminative process on each layer regardless that some of them remain frozen during training.

\section*{Acknowledgements}
\label{acknowledgments}

This work was funded by CONCYTEC-FONDECYT under the call E041-01 [contract number 34-2018-FONDECYT-BM-IADT-SE].

\bibliographystyle{coling}
\bibliography{semeval2020}

\end{document}